\documentclass{article}

\usepackage[preprint]{neurips_2026}

\usepackage[utf8]{inputenc} 
\usepackage[T1]{fontenc}    
\usepackage{hyperref}       
\usepackage{url}            
\usepackage{booktabs}       
\usepackage{amsfonts}       
\usepackage{nicefrac}       
\usepackage{microtype}      
\usepackage{xcolor}         
\usepackage{svg}
\usepackage{pifont}
\newcommand{\cmark}{\ding{51}} 
\newcommand{\xmark}{\ding{55}} 
\newcommand{\pmark}{\textbf{
\checkmark\kern-1.1ex\raisebox{.7ex}{\rotatebox[origin=c]{125}{--}}}}

\usepackage{graphicx}
\usepackage{wrapfig}
\usepackage{amsmath}
\usepackage{adjustbox}
\usepackage{multirow}
\usepackage{lipsum}
\usepackage{subcaption} 
\usepackage{amsfonts}       
\usepackage{xspace}         
\usepackage{wrapfig,lipsum}
\usepackage{colortbl}
\usepackage[capitalize]{cleveref}  
\crefname{section}{Sec.}{Secs.}
\Crefname{section}{Section}{Sections}
\crefname{table}{Tab.}{Tabs.}
\Crefname{table}{Table}{Tables}
\crefname{figure}{Fig.}{Figs.}
\Crefname{figure}{Figure}{Figures}
\crefname{equation}{Eq.}{Eqs.}
\Crefname{equation}{Equation}{Equations}
\crefname{appendix}{Appx.}{Appxs.}
\Crefname{appendix}{Appendix}{Appendices}
\newcommand{\method}{\texttt{OmniEdit-Bench}\xspace}
\title{OmniEdit-Bench: A Comprehensive Benchmark for Instruction-based Video Editing}

\author{%
  Chenxuan Miao$^1$, Yutong Feng$^2$\thanks{Project Lead}, Yi Lu$^4$, Yunfeng Yan$^3$, Donglian Qi$^3$, \\
  \textbf{Shiwei Zhang$^2$, Yu Liu$^2$, Xi Chen$^1$, Hengshuang Zhao$^1$}\thanks{Corresponding Author} \\
  \vspace{3pt}
  $^1$The University of Hong Kong $^2$Wan Team, Alibaba Group $^3$Zhejiang University $^4$Peking University \\
  \texttt{\{weiyuchoumou526, fengyutong.fyt\}@gmail.com} \\
  \texttt{hszhao@cs.hku.hk}
}

\begin{document}

\maketitle


\begin{abstract}
Instruction-based video editing (IVE) offers diverse applications, yet evaluating editing models remains challenging.
Existing benchmarks suffer from two key limitations: narrow task coverage inherited from image editing that overlooks video-specific dimensions, and metrics that inadequately capture instruction fidelity, allowing visually plausible but incorrect edits to receive high scores.
To address these limitations, we introduce \method, a comprehensive benchmark covering spatial, temporal, audio, and reference-based editing.
It distinguishes between explicit and implicit instructions and incorporates reasoning-based scenarios to reflect real-world editing requirements.
We further propose an evaluation framework that assesses accuracy, preservation, realism, and consistency using both human judges and a state-of-the-art vision-language model (VLM).
An accuracy-aware penalty mechanism conditions the other three scores on accuracy, prioritizing instruction fidelity and penalizing plausible but incorrect edits.
Experiments on prominent open-source and commercial IVE models reveal substantial room for improvement.
\method provides a structured testbed for identifying model limitations and advancing instruction-based video editing.
Project page: https://omniedit-bench.github.io.
              
\end{abstract}

\section{Introduction}
\begin{figure*}[t!]
    \centering
    \includegraphics[scale=0.4]{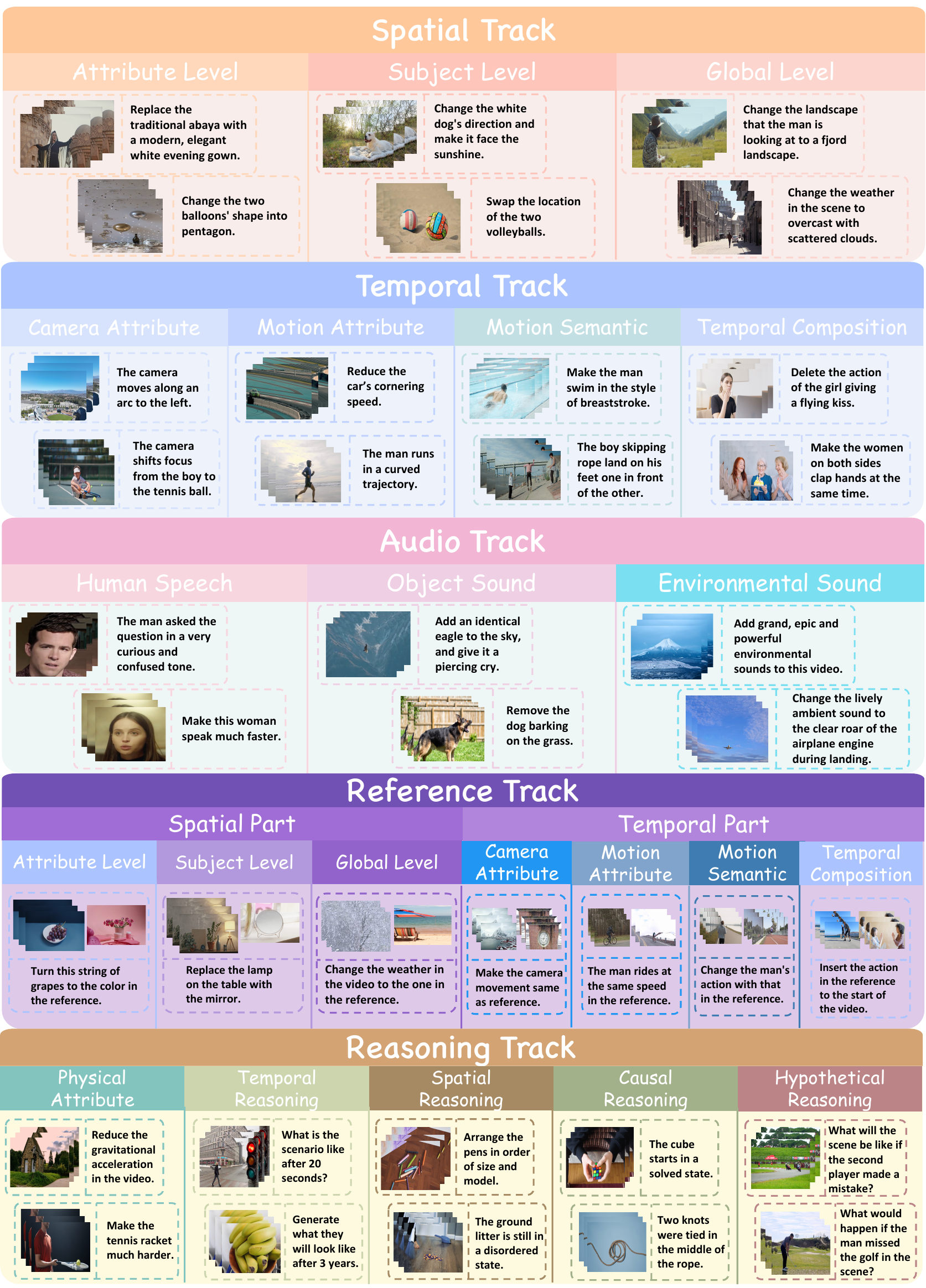}
    \caption{\textbf{Overview of our evaluation tracks.} This figure illustrates representative tasks from \textbf{Spatial}, \textbf{Temporal}, \textbf{Audio}, \textbf{Reference}, and \textbf{Reasoning} tracks. By covering a wide range of scenarios, our benchmarks provide a more comprehensive and precise basis for evaluating video editing models.}
    \vspace{-20pt}
    \label{fig:tracks_overview}
\end{figure*}
Instruction-based video editing (IVE) has emerged as a promising paradigm for controllable video manipulation. 
%
%
With the rapid advances in generative modeling~\cite{ho2020ddpm,song2020denoising,rombach2022high,song2021scorebased,yang2026geniedrive} and multimodal understanding~\cite{yang2024cogvideox}, instruction-based video editing has seen remarkable progress~\cite{zisenorita,bai2025ditto,cong2025viva,wei2025univideo}, enabling diverse operations such as object removal~\cite{miao2025rose,zhou2023propainter,li2025diffueraser,jiang2025smarteraser} and style transformation~\cite{isola2017pix2pix,zhu2017cyclegan,zhang2023inST}, and significantly advancing applications in video production.

Despite the rapid progress, evaluating IVE models remains a critical yet underexplored problem. 
%
%
The difficulty of evaluating instruction-based video editing stems from the inherently multi-dimensional nature of video content, which fundamentally distinguishes it from image editing~\cite{chen2024anydoor,chen2025unireal,ruiz2023dreambooth,brooks2023instructpix2pix,hertz2022prompt2prompt,zeng2025lumen}. 
Beyond spatial dimension, video editing involves additional dimensions such as temporal dynamics and audio signals.
%
%
However, existing benchmarks~\cite{chen2025ivebench,chen2025editboard,sun2025vebench,mou2025instructx} fail to adequately account for this multi-dimensional complexity.
Their task coverage remains narrow and largely inherited from image editing, focusing on frame-level spatial manipulations while neglecting video-specific dimensions such as temporal dynamics and audio~\cite{tan2024edtalk,oord2016wavenet,kong2020hifigan}. 
Moreover, current evaluation metrics~\cite{zhang2018lpips,wang2004ssim,psnr,huang2023vbench,hessel2021clipscore} are substantially misaligned with the core objective of instruction-based editing. 
Metrics adapted by current editing models fail to reflect instruction fidelity, allowing models to achieve high scores despite incorrect edits due to the strong visual prior of the original video.

To this end, we introduce \method, a comprehensive, well-categorized benchmark tailored for evaluating instruction-based video editing capabilities.
%
Our design explicitly accounts for the inherited multi-dimensional nature of video content, organizing tasks to reflect diverse editing requirements beyond static spatial changes. 
Specifically, we deliberately divides the benchmark into spatial, temporal, audio and reference-based track.
Each track captures a unique facet of how videos can be manipulated under instructions. 
Furthermore, we structure our benchmark along the axis of instruction complexity. 
Specifically, we distinguish between explicit instructions, which directly specify the desired edits, and implicit instructions, which require models to infer user intent from context.
Within the implicit setting, we include reasoning-based scenarios that involve multi-step or compositional transformations across one or multiple dimensions. 
By organizing along these two complementary axes, \method enables a more comprehensive and diagnostic evaluation.

For evaluation, we decompose the editing quality of videos into four dimensions: accuracy, preservation, realism, and consistency. 
This decomposition reflects the inherently multi-faceted nature of instruction-based video editing, where a successful edit must not only correctly execute the intended modification, but also preserve irrelevant content, maintain visual plausibility, and ensure coherence across frames and modalities.
Importantly, these dimensions are not equally informative in isolation. 
In particular, high visual preservation quality or temporal consistency does not necessarily imply correct execution of the intended edit. 
To address this, we enforce instruction fidelity as a prerequisite for meaningful evaluation by introducing an accuracy-aware scoring mechanism, in which the scores of other dimensions are conditioned on accuracy. 
Specifically, the scores of the remaining dimensions are proportionally scaled by the accuracy score, reducing their contributions when accuracy is low.
This design prevents visually plausible yet incorrect edits from receiving misleading high scores, leading to a more reliable and faithful assessment.

Using \method, we conduct a comprehensive evaluation of state-of-the-art instruction-based video editing models comprising both proprietary and open-source models. 
Our evaluation reveals consistent performance gaps across both open-source and commercial models.
All models perform poorly on video-specific tracks such as temporal and audio editing as well as on more challenging reasoning-based scenarios, highlighting substantial room for improvement in handling multi-dimensional and implicit instructions. 
Horizontally, on relatively well-studied tracks such as spatial track, commercial models such as KlingV3-Omni~\cite{team2025kling}, Runway Aleph~\cite{runway2024aleph} and Seedance2.0~\cite{seedance2026seedance}consistently achieve higher quantitative scores than open-source counterparts. This performance gap is also evident in qualitative aspects, including higher visual fidelity and longer temporal extent, reflecting advantages in data scale and system optimization.
\section{Related Work}
\begin{wrapfigure}[14]{r}{0.50\columnwidth}
\centering
\scriptsize
\setlength{\tabcolsep}{2.5pt}
\renewcommand{\arraystretch}{0.85}

\captionsetup{type=table}
\caption{Comparison of OmniEdit-Bench with existing benchmarks across different editing dimensions. 
\cmark\ indicates support, \xmark\ indicates lack of coverage, and \pmark\ indicates partial support.}
\label{tab:benchmark_comparison}

\begin{tabular}{@{}lccccc@{}}
\toprule
\textbf{Benchmark} 
& \textbf{Spatial} 
& \textbf{Temporal} 
& \textbf{Audio} 
& \textbf{Reference} 
& \textbf{Reasoning} \\
\midrule
VE-Bench~\cite{sun2025vebench}  & \cmark & \xmark & \xmark & \xmark & \xmark \\
EditBoard~\cite{chen2025editboard} & \cmark & \xmark & \xmark & \xmark & \xmark \\
OpenVE-3M~\cite{he2025openve}   & \cmark & \xmark & \xmark & \xmark & \xmark \\
IVE-Bench~\cite{chen2025ivebench} & \cmark & \pmark & \xmark & \xmark & \xmark \\
VIE-Bench~\cite{mou2025instructx} & \cmark & \xmark & \xmark & \cmark & \xmark \\
RVE-Bench~\cite{liu2025revise}   & \cmark & \xmark & \xmark & \xmark & \cmark \\
\midrule
\textbf{OmniEdit-Bench} 
& \cmark & \cmark & \cmark & \cmark & \cmark \\
\bottomrule
\end{tabular}
\end{wrapfigure}
\vspace{-6pt}
\textbf{Instruction-based Video Editing.}
Instruction-based video editing (IVE) has recently emerged as a compelling paradigm. Early approaches~\cite{guo2023animatediff} largely adapt image editing methods to the temporal domain by incorporating attention or consistency modules into text-to-image diffusion frameworks~\cite{saharia2022imagen,peebles2023dit,labs2025flux}. While such extensions can preserve spatial structure, they remain limited in handling dynamic scenarios involving object motion or camera movement. More recent methods adopt architectures including video diffusion~\cite{ho2022vdm} and flow-matching models~\cite{lipman2023flowmatching,liu2022rectifiedflow}, which improve temporal coherence and editing fidelity. Representative open-source systems include Señorita-2M~\cite{zisenorita}, Ditto~\cite{bai2025ditto}, VIVA~\cite{cong2025viva}, and UniVideo~\cite{wei2025univideo}, while commercial models such as Kling Omni~\cite{team2025kling}, Seedance2.0~\cite{seedance2026seedance}, Wan~\cite{wan2025wan,cheng2025wananimate}, and Grok Imagine~\cite{xai2024grokimagine} have achieved strong empirical performance. Despite these advances, model capabilities remain uneven, with systems excelling on different editing types, underscoring the need for systematic and standardized evaluation.

\textbf{Benchmarks for Instruction-based Video Editing.}
In parallel with the development of IVE methods, several evaluation benchmarks have been proposed. Early efforts such as VE-Bench~\cite{sun2025vebench} and EditBoard~\cite{chen2025editboard} focus on a limited set of tasks largely inherited from image editing, and remain constrained in both scale and diversity. Later benchmarks, including OpenVE-3M~\cite{he2025openve} and IVE-Bench~\cite{chen2025ivebench}, expand the data coverage but do not explicitly capture video-specific aspects such as temporal and audio. Some works further explore individual dimensions, where VIE-Bench~\cite{mou2025instructx} studies reference-conditioned editing, and RISE-Bench~\cite{zhao2025risebench} together with RVE-Bench~\cite{liu2025revise} investigate reasoning-based editing. However, these efforts remain fragmented and do not provide a unified evaluation framework. To address this gap, we propose \method, a comprehensive and structured benchmark for multi-dimensional evaluation of instruction-based video editing.
\section{OmniEdit-Bench}
\begin{figure*}[t!]
    \centering
    \includegraphics[width=0.75\linewidth]{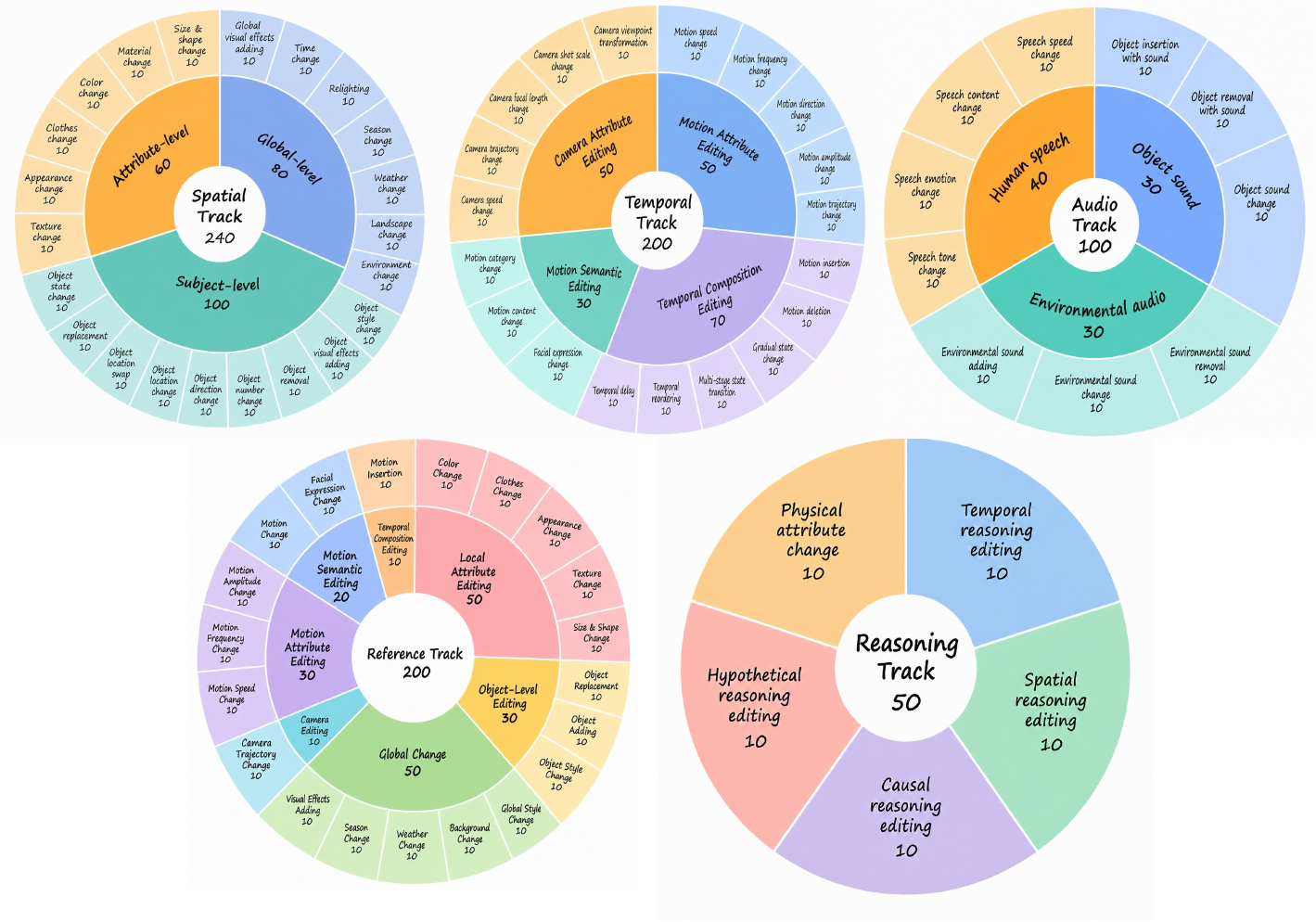}
    \caption{\textbf{Taxonomy of video editing tasks.} We classify the benchmark into five tracks: Spatial (240), Temporal (200), Reference (200), Audio (100), and Reasoning (50), illustrating the diverse scope of instruction-based video manipulation.}
    \vspace{-10pt}
    \label{fig:tasks_distribution}
\end{figure*}
\subsection{Benchmark Construction}
To comprehensively evaluate IVE task, we construct \method as a multi-dimensional and systematically organized benchmark. 
As illustrated in \cref{fig:tracks_overview} and \cref{fig:tasks_distribution}, our design decomposes editing tasks along complementary axes, covering explicit and implicit instruction settings as well as diverse video-specific dimensions. Specifically, the benchmark includes five tracks: \textbf{spatial}, \textbf{temporal}, \textbf{audio}, \textbf{reference-based}, and \textbf{reasoning}, each targeting a distinct aspect of editing capability. 
This design enables fine-grained and diagnostic evaluation of models across different levels of complexity, ranging from low-level visual manipulation to high-level reasoning and multimodal alignment.

\textbf{Spatial Track.}
The spatial track focuses on scenarios with limited motion, where consecutive frames remain highly similar. 
This setting largely aligns with image editing~\cite{chen2024anydoor,chen2025unireal} and is designed to evaluate models’ ability to perform fine-grained visual modifications without substantial temporal variation. 
To systematically characterize different levels of editing granularity, we further divide this track into three sub-tracks: attribute-level, subject-level, and global-level editing. 

\textit{Attribute-level editing} targets low-level fine-grained changes, such as color, texture, or material, and is designed to evaluate whether models can precisely modify local visual properties.

\textit{Subject-level editing} focuses on object-centric changes, including removal, replacement, spatial manipulations or style changes . This category is defined by an intermediate level of granularity, where the scope of editing expands from local attribute changes to semantically entire objects. 

\textit{Global-level editing} involves scene-wide transformations, such as relighting, weather or season changes, and represents the coarsest level of granularity. This category evaluates whether models can perform coherent transformations that affect the scene as a whole rather than isolated regions.
\begin{figure*}[t!]
    \centering
    \includegraphics[width=0.8\linewidth]{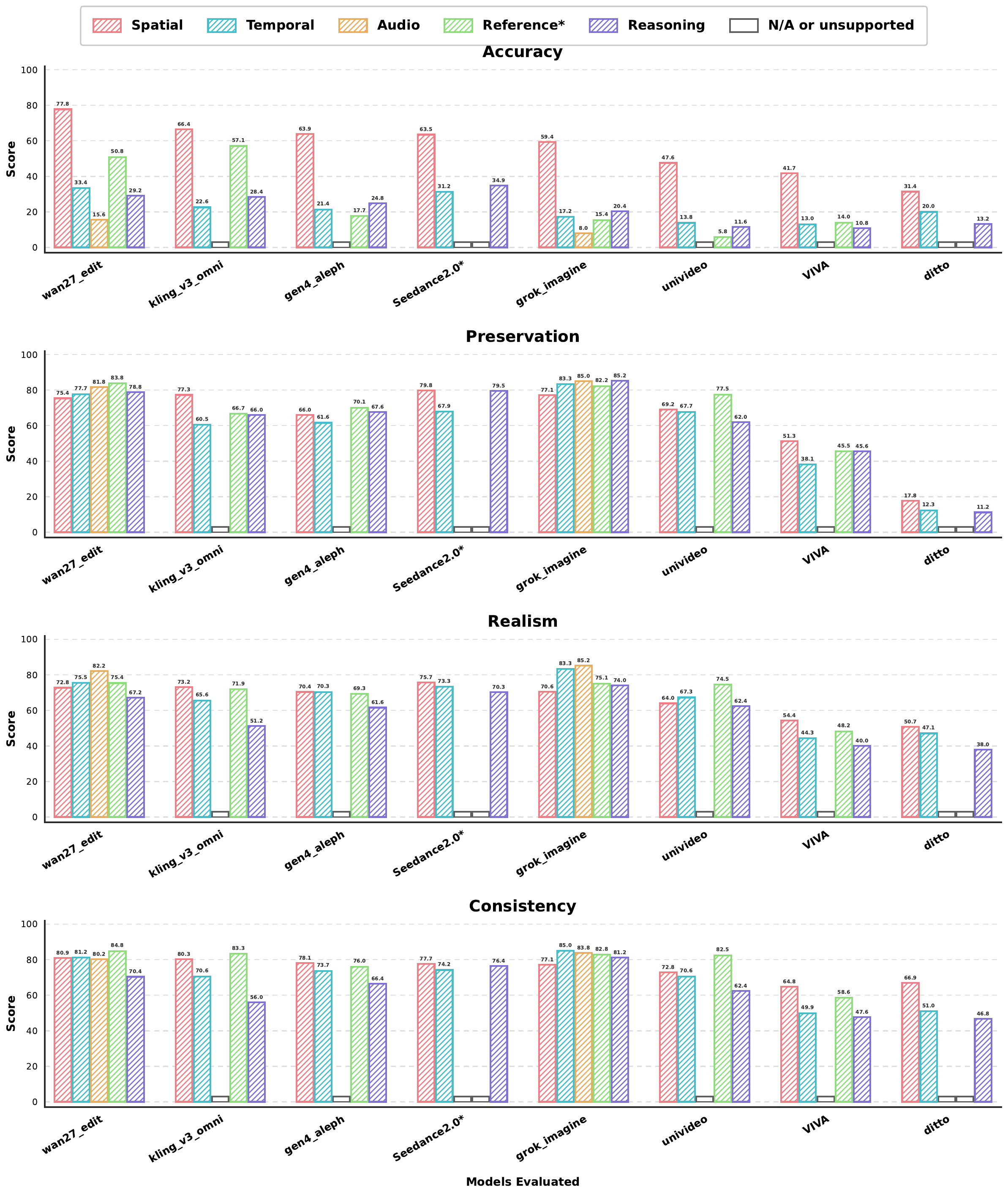}
    \caption{\textbf{Comparisons of model performance across evaluation metrics.} The bar charts display the score distribution of all the evaluated models for Accuracy, Preservation, Realism, and Consistency, categorized by the five tracks defined in OmniEdit-Bench. \textit{Note: For Reference track, only models with spatial reference capabilities are evaluated due to current architectural limitations.}}
    \vspace{-22pt}
    \label{fig:dimension_wise_results}
\end{figure*}
\textbf{Temporal Track.}
Temporal track is designed to capture the intrinsic temporal complexity of video editing, which is mostly absent in previous benchmarks. 
This track focuses on videos with significant camera or object motion and we organize this track into four sub-tracks: camera attribute editing, motion attribute editing, motion semantic editing, and temporal composition editing. 

\textit{Camera attribute editing} targets transformations in camera parameters, such as viewpoint, shot scale, and trajectory. These operations require modeling continuous camera motion over time, introducing long-range temporal dependencies across frames. Models must ensure smooth transitions, consistent motion trajectories, and stable geometric relationships throughout the video. 
\begin{figure*}[t!]
    \centering
    \includegraphics[width=0.95\linewidth]{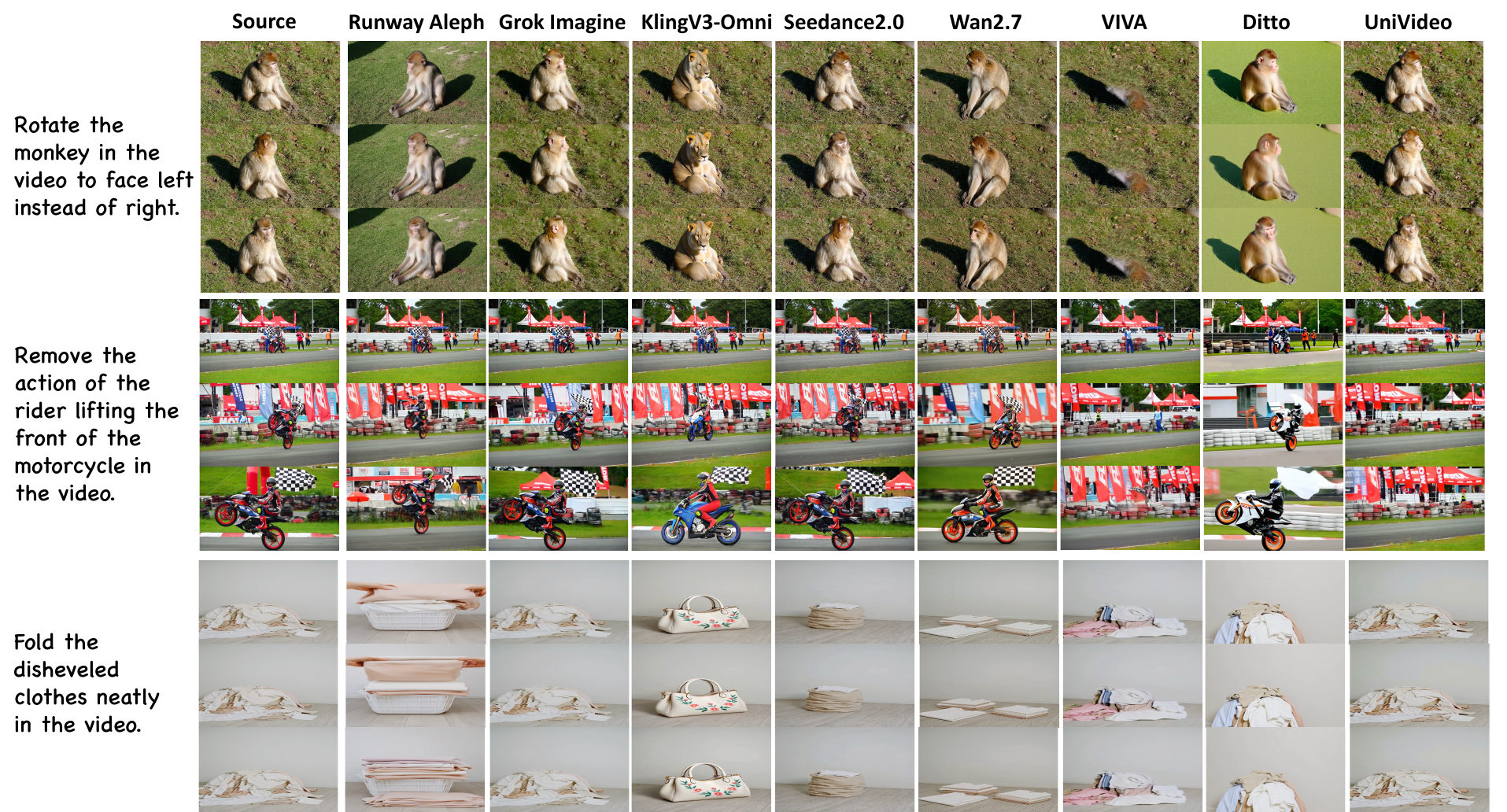}
    \caption{\textbf{Some examples of models' output on \method.} Zoom in for better view. For more visualization results, please refer to \cref{appd:visualization_results}.}
    \label{fig:visualizations}
    \vspace{-18pt}
\end{figure*}
\textit{Motion attribute editing} focuses on controlling quantitative and basic properties of motion, such as speed, direction, and amplitude. Unlike camera attribute editing modifying the observation perspective, this category directly alters object dynamics within the scene. This sub-track evaluates whether models can precisely modulate motion signals while maintaining coherent temporal evolution.
%
\textit{Motion semantic editing} involves higher-level transformations of motion content or category, such as altering the type or meaning of an action. This category requires models to understand and recompose semantically meaningful motion patterns over time. Such edits typically span multiple frames and involve structured temporal dependencies, where the global motion pattern must remain temporally coherent while being semantically transformed. This sub-track evaluates whether models can perform temporally consistent motion reinterpretation rather than merely adjusting low-level motion statistics.

\textit{Temporal composition editing} captures more complex operations over entire motion sequences, including motion insertion, reordering, and synchronization. These operations introduce long-range temporal dependencies and often involve multi-step transformations, where local consistency must be preserved while enforcing new temporal relationships across distant frames. This makes temporal composition editing particularly challenging, as models must coordinate motion events over the video and avoid inconsistencies such as temporal misalignment or incoherent transitions. 

\textbf{Audio Track.}
Video inherently involves visual and auditory signals, making audio a critical yet underexplored dimension for IVE task.
Audio track evaluates whether models can perform coherent audio-visual editing, where modifications to the visual content must be accompanied by temporally aligned and semantically consistent changes in audio.
This track is divided into three sub-tracks: human speech editing, object sound editing, and environmental audio editing. 

\textit{Human speech editing} focuses on modifications related to human speech, including changes in speech content, speaking speed, emotion, and tone. This category evaluates whether models can generate temporally aligned speech signals that are consistent with the visual context, such as lip movements and speaker identity, while preserving natural prosody and semantic correctness. 

\textit{Object sound editing} addresses audio changes associated with object-level interactions. Typical tasks include inserting or removing objects along with their corresponding sounds, or modifying the sounds produced by objects. This category focuses more on general objects inluding animals and vehicles and requires models to maintain consistency between visual events and their associated sounds.

\textit{Environmental audio editing} focuses on background or ambient sounds, such as adding, modifying, or removing environmental audio. These edits require maintaining global audio consistency over time while ensuring compatibility with the visual scene and the corresponding editing instruction.

\textbf{Reference-based Track.} 
We further introduce a reference-based track to evaluate models’ ability to incorporate external visual guidance during editing.
Rather than defining a completely independent set of tasks, this track is built upon both spatial and temporal editing scenarios by introducing reference inputs, requiring models to perform edits that are not only instruction-following but also aligned with a given reference.
In the spatial setting, reference-based tasks are instantiated using both reference images and reference videos, where both forms primarily provide spatial guidance.
The temporal setting relies exclusively on reference videos, as dynamic transformations such as motion patterns and camera trajectories can not be adequately specified by a single image. 
%
%
We further organize reference-based tasks following the structural taxonomy as spatial and temporal editing, enabling systematic evaluation of whether models can not only perform edits but also faithfully align results with reference signals across both spatial and temporal dimensions.

\textbf{Reasoning Track.}
Reasoning track is designed with using implicit editing instructions where the desired editing outcome is not directly specified but must be inferred from the instruction.
%
These tasks require models to interpret user intent and perform multi-step reasoning before generating the edited video. 
We organize these tasks into five categories, each capturing a distinct type of reasoning.

\textit{Physical attribute change} involves modifying underlying physical properties in the scene, such as gravity, friction, or weight, requiring models to reason about physically plausible outcomes over time.
\begin{figure*}[ht!]
    \centering
    \includegraphics[width=0.8\linewidth]{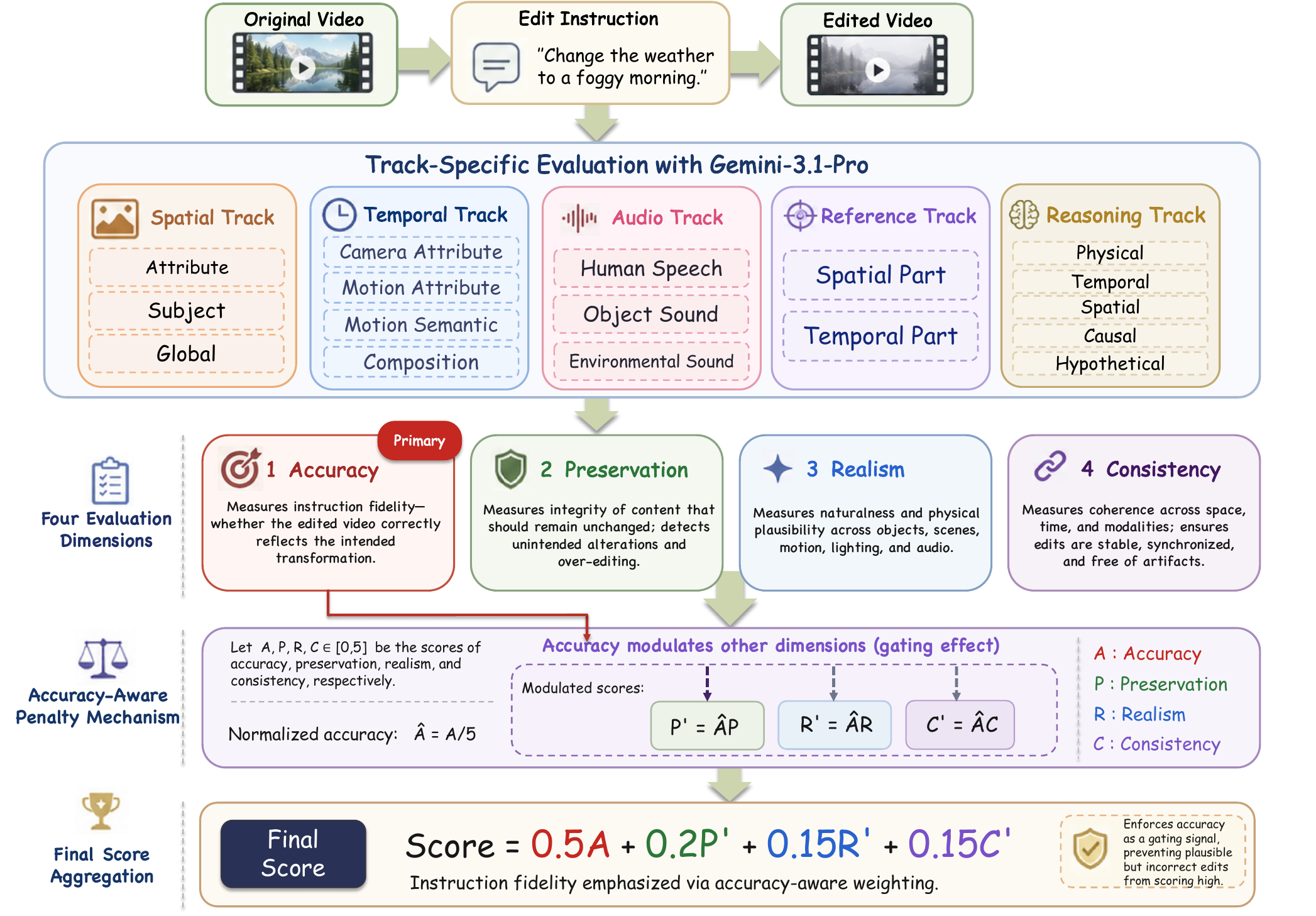}
    \caption{\textbf{Evaluation framework.} By introducing an accuracy-aware penalty mechanism, our framework ensures a more precise assessment of video editing quality. The accuracy score acts as a gating signal, preventing plausible-looking but instruction-deviant edits from receiving high scores.}
    \label{fig:evaluation_pipeline}
    \vspace{-10pt}
\end{figure*}
\textit{Spatial reasoning editing} focuses on inferring spatial arrangements and relationships, such as reorganizing a cluttered scene into a structured layout, demanding an understanding of object grouping, alignment, and spatial constraints.

\textit{Temporal reasoning editing} requires models to infer time-dependent outcomes, such as predicting scene evolution after a given duration (e.g., traffic light changes after a certain time), emphasizing temporal progression and consistency.

\textit{Causal reasoning editing} involves reasoning about cause-and-effect relationships, such as determining the state of a system given certain conditions (e.g., a solved Rubik’s cube), requiring models to generate outcomes consistent with underlying causal logic.

\textit{Hypothetical reasoning editing} addresses counterfactual or imaginative scenarios (e.g., “what if” conditions), where models must generate plausible alternative outcomes that are not directly observed, requiring abstraction beyond the given input.
\subsection{Evaluation Pipeline}
To systematically evaluate instruction-based video editing performance, as shown in \cref{fig:evaluation_pipeline}, we design a multi-dimensional evaluation pipeline that decomposes editing quality into four complementary aspects: \textbf{accuracy}, \textbf{preservation}, \textbf{realism}, and \textbf{consistency}. 
Each dimension is independently assessed using a state-of-the-art vision-language model (VLM), specifically Gemini-3.1-Pro, with task-specific prompts tailored to different tracks (spatial, temporal, audio, reference, and reasoning). The detailed prompt designs for each track are illustrated in \cref{appd:prompts_for_vlm}.

\textbf{Accuracy.}
Accuracy measures whether the edited video correctly reflects the intended transformation specified by the instruction. 
It serves as the primary indicator of instruction fidelity.
The evaluation of accuracy varies across tracks to capture task-specific requirements. 
For spatial editing, accuracy is evaluated at different levels of granularity.
%
For temporal editing, it further includes action content accuracy, motion attribute accuracy, timing accuracy, and camera control accuracy, reflecting both what is edited and when it occurs. 
For audio editing, accuracy considers both sound correctness and temporal alignment. 
In the reasoning track, accuracy evaluates whether the model successfully infers and executes implicit instructions, including spatial, temporal, and causal reasoning.
\textbf{Preservation.}
Preservation evaluates whether the edited video maintains the integrity of content that should remain unchanged. 
In IVE task, only part of the scene is expected to be modified, and unintended alterations to irrelevant regions indicate poor editing control.
This dimension is particularly important for distinguishing precise editing from over-editing. 
%
%

\begin{table*}[!ht]
    \centering
    \scriptsize 
    \renewcommand{\arraystretch}{1} 
    
    \caption{\textbf{Overall performance on OmniEdit-Bench.} We report the performance of various video editing models across five primary tracks. Higher scores indicate better alignment with the evaluation dimensions. 
    \textit{Note: Seedance 2.0 is only evaluated on Spatial, Temporal, and Reasoning tracks due to copyright restrictions. And for Reference track, only models with spatial reference capabilities are evaluated due to current architectural limitations.}}
    \label{tab:overall_performance}
    \begin{tabular}{lccccccc}
    \toprule
    \textbf{Models} & \textbf{Resolution} & \textbf{Spatial} & \textbf{Temporal} & \textbf{Audio} & \textbf{Reference$^{*}$} & \textbf{Reasoning} & \textbf{Overall Score (Average)}\\
    \midrule
    Runway Aleph~\citep{runway2024aleph} & 720P & 54.2 & 16.5  & - & 15.1  & 20.3 & 26.5  \\
    Grok Imagine~\citep{xai2024grokimagine} & 720P & 51.7 & 15.0 & 6.7  & 13.1 & 17.9  & 20.9 \\
    KlingV3-Omni~\cite{team2025kling} & 720P & 59.6  & 18.5  & - & 49.9 & 22.9  & 37.7  \\
    Seedance2.0$^{*}$~\citep{seedance2026seedance}  & 720P & 56.9  & 25.2  & -  & -  & 29.8  &  37.3 \\
    Wan2.7-Edit~\citep{wan2025wan}  & 720P & 69.8  & 29.0  & 13.6  & 46.4 & 24.6 & 36.7  \\
    VIVA~\citep{cong2025viva}        & 480P & 33.5  & 9.6 & - & 9.8  & 7.5 & 15.1 \\
    Ditto~\citep{bai2025ditto}        & 480P & 22.9  & 13.5 & - & -  & 8.6  & 15.0  \\
    UniVideo~\citep{wei2025univideo}     & 480P & 39.9  & 11.3 & - & 4.9  & 8.4  & 16.1  \\
    \bottomrule
    \end{tabular}
\end{table*}
\textbf{Realism.}
Realism measures whether the edited video appears natural and physically plausible. This dimension captures perceptual quality beyond instruction correctness.
We evaluate realism at multiple levels. 
Physical realism assesses whether the edited content obeys physical constraints, such as lighting consistency and shadows.
Subject realism examines whether the appearance of objects or characters remains plausible after editing.
Environment realism evaluates the coherence of the overall scene. 
For temporal and motion-related tasks, we additionally consider motion realism, ensuring that dynamic behaviors (e.g., speed, frequency, trajectory) appear natural rather than artificial.
For audio-related tasks, realism also includes the naturalness of generated or modified audio signals.

\textbf{Consistency.}
Consistency evaluates whether the editing results are coherent. 
We consider three forms of consistency. 
Temporal consistency ensures that edits are smoothly and coherently applied across frames without flickering or discontinuities. 
Spatial consistency evaluates whether spatial relationships and structures remain stable after editing. For audio-related tasks, we additionally consider audio-visual consistency, ensuring that audio signals are synchronized with visual content.

\textbf{Accuracy-Aware Penalty Mechanism.}
To emphasize instruction fidelity, as illustrated in \cref{fig:evaluation_pipeline}, we introduce an accuracy-aware penalty mechanism that adjusts the contributions of different evaluation dimensions. 
Let $A, P, R, C \in [0,5]$ denote the scores of accuracy, preservation, realism, and consistency, respectively, and define the normalized accuracy as $\hat{A} = A/5$. 
The scores of the remaining dimensions are proportionally modulated as $P’ = \hat{A}P$, $R’ = \hat{A}R$, and $C’ = \hat{A}C$. 
The final score is then computed as a weighted combination: $\text{Score} = 0.5A + 0.2P’ + 0.15R’ + 0.15C’$. 
In this formulation, accuracy serves as a controlling factor that continuously scales the influence of other dimensions, ensuring that their contributions diminish when the editing accuracy is low. 
This design prevents visually plausible but incorrect edits from receiving inflated scores, while still allowing high-quality edits to benefit from strong performance across all dimensions.
\vspace{-5pt}
\section{Experiments}
\subsection{Main Results}
We evaluate a diverse set of state-of-the-art instruction-based video editing models on \method, covering both open-source and commercial systems. 
The results are summarized in \cref{tab:overall_performance}, where we report performance on a 100-point scale across five tracks.
To further analyze model behavior beyond the aggregated scores, we report dimension-wise evaluation results in \cref{fig:dimension_wise_results} across different tracks and models.
In addition, to validate the reliability of our evaluation protocol, we compare VLM-based scores with human annotations, as shown in \cref{tab:vlm_human_alignment}.
\subsection{Comparison across Models}
\cref{tab:overall_performance} reveals clear performance differences across models. 
Overall, commercial systems consistently outperform open-source models, particularly on well-studied tasks such as spatial editing. 
%
%
However, this advantage diminishes on more challenging tracks.
Even the strongest models show limited performance on temporal and reasoning tasks. 
For example, while Wan2.7-Edit reaches 29.0 on the temporal track, most other models remain below 20, and reasoning performance remains modest overall, with the best score (Seedance2.0) only reaching 29.8. 
This suggests that improvements in model scale and training data primarily benefit spatial tasks, but are insufficient for addressing more complex temporal and reasoning challenges.
On the reference track, performance varies significantly across models, with KlingV3-Omni (49.9) and Wan2.7-Edit (46.4) outperforming others by a large margin, reflecting stronger capabilities in reference-conditioned generation.
Open-source models, while generally lagging behind, still exhibit competitive performance in certain settings. 
For example, UniVideo achieves a reasonable spatial score (39.9), indicating potential for further improvement with better training strategies and architectural design.
\subsection{Comparisons across Tracks}
From a task perspective, we observe consistent performance disparities across different tracks. 
While models achieve relatively strong results on the spatial track (e.g., up to 69.8), their performance drops significantly on temporal, audio, and reasoning tracks.
The temporal track remains particularly challenging, with most models scoring below 20 and only a few exceeding 25 (e.g., Seedance2.0 at 25.2 and Wan2.7-Edit at 29.0). 
This reflects the difficulty of maintaining coherent motion and modeling long-range temporal dependencies across frames.
The reasoning track further exposes limitations in instruction understanding. 
Despite being a core component of realistic editing scenarios, reasoning scores remain relatively low across all models (mostly below 25), indicating that current systems struggle to infer implicit instructions.
The audio track highlights a critical gap in multimodal capabilities. 
Only a subset of models supports audio editing, and overall performance remains low, confirming that audio remains an underexplored dimension in video editing.
These results suggest that while current models perform relatively well on spatial editing, they remain far from handling realistic video editing scenarios that require temporal coherence, multimodal alignment, and reasoning ability, which points the way for future research in IVE task.
\subsection{Human Alignment}
\begin{table*}[t]
\centering
\small
\setlength{\tabcolsep}{4pt}
\caption{\textbf{Alignment between VLM-based evaluation and human judgments.} 
For each VLM score level (0--5), we report the proportion of samples, 
the mean and standard deviation of human scores, and the mean absolute error (MAE), 
across four evaluation dimensions: Accuracy (Acc.), Preservation (Pres.), 
Realism (Real.), and Consistency (Cons.).}
\label{tab:vlm_human_alignment}

\resizebox{\textwidth}{!}{
\begin{tabular}{c|cccc|cccc|cccc|cccc}
\toprule
\multirow{2}{*}{Score} 
& \multicolumn{4}{c|}{Prop. (\%)} 
& \multicolumn{4}{c|}{Human Mean} 
& \multicolumn{4}{c|}{Human Std.} 
& \multicolumn{4}{c}{MAE} \\
\cmidrule(lr){2-5} \cmidrule(lr){6-9} \cmidrule(lr){10-13} \cmidrule(lr){14-17}
& Acc. & Pres. & Real. & Cons. 
& Acc. & Pres. & Real. & Cons. 
& Acc. & Pres. & Real. & Cons. 
& Acc. & Pres. & Real. & Cons. \\
\midrule
0 & 21.98 & 12.10 & 10.12 & 8.37 & 0.23 & 0.18 & 0.31 & 0.21 & 0.08 & 0.07 & 0.13 & 0.10 & 0.23 & 0.18 & 0.31 & 0.21\\
1 & 19.37 & 9.97 & 6.72 & 9.10 & 1.48 & 1.80 & 1.28 & 1.89 & 0.55 & 0.42 & 1.20 & 1.40 & 0.85 & 0.83 & 0.80 & 1.61 \\
2 & 13.64 & 9.88 & 13.91 & 10.77 & 2.18 & 2.64 & 2.37 & 2.15 & 1.50 & 1.22 & 1.10 & 1.24 & 1.64 & 0.96 & 0.84 & 0.75 \\
3 & 5.52 & 9.86 & 15.34 & 10.18 & 3.87 & 3.33 & 3.73 & 3.50 & 0.83 & 0.96 & 0.80 & 0.78 & 0.77 & 0.70 & 0.60 & 0.78 \\
4 & 14.02 & 16.10 & 23.12 & 24.68 & 4.26 & 3.81 & 3.67 & 4.46 & 0.94 & 1.46 & 0.64 & 0.83 & 0.70 & 1.12 & 0.76 & 0.79 \\
5 & 25.47 & 42.09 & 30.78 & 36.90 & 4.61 & 4.44 & 4.57 & 4.87 & 1.12 & 0.88 & 0.51 & 0.34 & 0.39 & 0.56 & 0.43 & 0.13 \\
\midrule
Overall & - & - & - & - & - & - & - & - & - & - & - & -
& \textbf{0.67} & \textbf{0.68} & \textbf{0.59} & \textbf{0.56} \\
\bottomrule
\end{tabular}
}
\end{table*}
\begin{table*}[t]
    \centering
    \small

    \begin{minipage}[t]{\dimexpr(\textwidth-\columnsep)/2\relax}
        \centering
        \caption{Ablation of accuracy-aware scoring.
        EA: Equal Average; WA: Weighted Average;
        AAP: Accuracy-aware Penalty.}
        \label{tab:accuracy_aware_ablation}
        
        \vspace{2pt}
        \footnotesize
        \renewcommand{\arraystretch}{1.15}
        \begin{tabular*}{0.95\linewidth}
            {@{\extracolsep{\fill}}lccc@{}}
            \toprule
            \textbf{Method}
            & \textbf{Human Corr.$\uparrow$}
            & \textbf{Pearson$\uparrow$}
            & \textbf{MAE$\downarrow$} \\
            \midrule
            EA & 0.846 & 0.861 & 0.468 \\
            WA & 0.889 & 0.894 & 0.392 \\
            \textbf{AAP}
               & \textbf{0.934}
               & \textbf{0.921}
               & \textbf{0.287} \\
            \bottomrule
        \end{tabular*}
    \end{minipage}
    \hfill
    \begin{minipage}[t]{\dimexpr(\textwidth-\columnsep)/2\relax}
        \centering
        \caption{Sensitivity analysis of different weight configurations.}
        \label{tab:weight_sensitivity}
        
        \vspace{2pt}
        \scriptsize
        \setlength{\tabcolsep}{3pt}
        \renewcommand{\arraystretch}{1.05}
        \begin{tabular}{@{}ccccccc@{}}
            \toprule
            $w_A$ & $w_P$ & $w_R$ & $w_C$
            & \textbf{Human $\rho\uparrow$}
            & \textbf{MAE$\downarrow$}
            & \textbf{Ranking $\tau\uparrow$} \\
            \midrule
            0.25 & 0.25 & 0.25  & 0.25
                 & 0.881 & 0.374 & 0.91 \\
            0.35 & 0.25 & 0.20  & 0.20
                 & 0.907 & 0.341 & 0.94 \\
            0.50 & 0.25 & 0.125 & 0.125
                 & 0.928 & 0.301 & \textbf{0.98} \\
            \textbf{0.50} & \textbf{0.20}
                 & \textbf{0.15} & \textbf{0.15}
                 & \textbf{0.934}
                 & \textbf{0.287}
                 & \textbf{0.98} \\
            0.50 & 0.15 & 0.25 & 0.10
                 & 0.919 & 0.315 & 0.96 \\
            0.60 & 0.16 & 0.12 & 0.12
                 & 0.925 & 0.306 & \textbf{0.98} \\
            0.70 & 0.12 & 0.09 & 0.09
                 & 0.910 & 0.329 & 0.94 \\
            \bottomrule
        \end{tabular}
    \end{minipage}

    \vspace{-5pt}
\end{table*}
To evaluate the reliability of evaluation pipeline, we compare VLM-based scores with human scores, as shown in \cref{tab:vlm_human_alignment}. 
Specifically, for each evaluation dimension, we group samples according to the discrete VLM scores (0--5) assigned by Gemini~3.1-pro, and report the proportion of samples in each score level. 
We then compute the corresponding human score statistics, including the mean, standard deviation, and mean absolute error (MAE) between VLM and human scores.
The results indicate strong alignment between VLM and human evaluation across all four dimensions.
Human mean scores closely track the VLM score levels, exhibiting an overall positive association.
For instance, samples assigned the highest VLM score (5) receive consistently high human ratings (4.44--4.87), while lower scores correspond to proportionally lower human evaluations, suggesting good calibration of the VLM.
The MAE remains relatively low across dimensions, with overall values of 0.86 (accuracy), 0.77 (preservation), 0.55 (realism), and 0.64 (consistency). 
This indicates that VLM evaluations are generally consistent with human assessments. 
Meanwhile, human score variance is moderate, with lower dispersion at higher score levels, implying stronger agreement on high-quality edits and greater ambiguity for intermediate cases.
These results demonstrate that VLM-based evaluation provides a reliable and scalable proxy for human judgment, supporting its use in large-scale benchmarking.

\subsection{Ablation and Sensitivity Analysis of Accuracy-aware Scoring}
\textbf{Effectiveness of Accuracy-aware Scoring.}
We first ablate the proposed accuracy-aware mechanism by comparing it with 
Equal Average and Weighted Average. 
As shown in ~\cref{tab:accuracy_aware_ablation}, introducing dimension-specific 
weights improves Human Corr. from 0.846 to 0.889 and reduces MAE from 
0.468 to 0.392. 
More importantly, incorporating the accuracy-aware penalty 
further improves Human Corr. to 0.934 and Pearson correlation to 
0.921, while reducing MAE to 0.287. 
This demonstrates that the improvement does not merely come from assigning 
a larger weight to Accuracy; explicitly penalizing inaccurate edits better 
captures its prerequisite role in successful editing.

\textbf{Sensitivity to Dimension Weights.}
We further analyze the sensitivity to different weight configurations in 
~\cref{tab:weight_sensitivity}. 
Increasing the emphasis on Accuracy initially improves human alignment, with our adopted configuration $(0.50, 0.20, 0.15, 0.15)$ achieving the highest Human Spearman correlation (0.934), the lowest MAE (0.287), and a high ranking consistency ($\tau=0.98$). However, further increasing the Accuracy weight degrades performance, indicating that Accuracy should be dominant 
rather than exclusive. 
These results support our design of treating Accuracy as the primary criterion while retaining Preservation, Realism, and Consistency as complementary dimensions.
\section{Conclusion}
\vspace{-5pt}
In this paper, we introduce \method, a comprehensive and systematically structured benchmark for instruction-based video editing, covering five tracks: spatial, temporal, audio, reference-conditioned, and reasoning. We further propose a unified evaluation pipeline that decomposes editing quality into four dimensions: accuracy, preservation, realism, and consistency, together with an accuracy aware penalty mechanism to ensure faithful assessment of instruction fidelity. Extensive experiments show that while existing models perform well on spatial editing, their performance degrades substantially on more challenging tracks involving temporal coherence, multimodal alignment, and implicit reasoning, with even state-of-the-art commercial models exhibiting clear limitations. These findings highlight fundamental challenges in instruction-based video editing and point to important future directions. 

\newpage
\small{\bibliographystyle{plain}
\bibliography{references}}

@inproceedings{ho2020ddpm,
  author    = {Jonathan Ho and Ajay Jain and Pieter Abbeel},
  title     = {Denoising Diffusion Probabilistic Models},
  booktitle = {{NeurIPS}},
  year      = {2020},
}

@article{song2020denoising,
  title={Denoising diffusion implicit models},
  author={Song, Jiaming and Meng, Chenlin and Ermon, Stefano},
  journal={arXiv preprint arXiv:2010.02502},
  year={2020}
}

@inproceedings{rombach2022high,
  title   = {High-Resolution Image Synthesis with Latent Diffusion Models},
  author  = {Robin Rombach and Andreas Blattmann and Dominik Lorenz and Patrick Esser and Björn Ommer},
  booktitle = {{CVPR}},
  year    = {2022}
}

@inproceedings{song2021scorebased,
  author    = {Yang Song and Jascha Sohl-Dickstein and Diederik P. Kingma and Abhishek Kumar and Stefano Ermon and Ben Poole},
  title     = {Score-Based Generative Modeling through Stochastic Differential Equations},
  booktitle = {{ICLR}},
  year      = {2021}
}

@article{yang2024cogvideox,
  title={Cogvideox: Text-to-video diffusion models with an expert transformer},
  author={Yang, Zhuoyi and Teng, Jiayan and Zheng, Wendi and Ding, Ming and Huang, Shiyu and Xu, Jiazheng and Yang, Yuanming and Hong, Wenyi and Zhang, Xiaohan and Feng, Guanyu and others},
  journal={arXiv preprint arXiv:2408.06072},
  year={2024}
}

@inproceedings{zisenorita,
  title={Se{\~n}orita-2M: A High-Quality Instruction-based Dataset for General Video Editing by Video Specialists},
  author={Zi, Bojia and Ruan, Penghui and Chen, Marco and Qi, Xianbiao and Hao, Shaozhe and Zhao, Shihao and Huang, Youze and Liang, Bin and Xiao, Rong and Wong, Kam-Fai},
  booktitle={{NeurIPS}},
  year={2025}
}

@article{bai2025ditto,
  title={Scaling instruction-based video editing with a high-quality synthetic dataset},
  author={Bai, Qingyan and Wang, Qiuyu and Ouyang, Hao and Yu, Yue and Wang, Hanlin and Wang, Wen and Cheng, Ka Leong and Ma, Shuailei and Zeng, Yanhong and Liu, Zichen and others},
  journal={arXiv preprint arXiv:2510.15742},
  year={2025}
}

@article{cong2025viva,
  title={VIVA: VLM-Guided Instruction-Based Video Editing with Reward Optimization},
  author={Cong, Xiaoyan and Yang, Haotian and Wang, Angtian and Wang, Yizhi and Yang, Yiding and Zhang, Canyu and Ma, Chongyang},
  journal={arXiv preprint arXiv:2512.16906},
  year={2025}
}

@article{wei2025univideo,
  title={Univideo: Unified understanding, generation, and editing for videos},
  author={Wei, Cong and Liu, Quande and Ye, Zixuan and Wang, Qiulin and Wang, Xintao and Wan, Pengfei and Gai, Kun and Chen, Wenhu},
  journal={arXiv preprint arXiv:2510.08377},
  year={2025}
}

@inproceedings{miao2025rose,
  title={ROSE: Remove Objects with Side Effects in Videos},
  author={Miao, Chenxuan and Feng, Yutong and Zeng, Jianshu and Gao, Zixiang and Hantang, Liu and Yan, Yunfeng and Qi, Donglian and Chen, Xi and Wang, Bin and Zhao, Hengshuang},
  booktitle={{NeurIPS}},
  year={2025}
}

@inproceedings{zhou2023propainter,
  title={Propainter: Improving propagation and transformer for video inpainting},
  author={Zhou, Shangchen and Li, Chongyi and Chan, Kelvin CK and Loy, Chen Change},
  booktitle={{ICCV}},
  year={2023}
}

@article{li2025diffueraser,
  title={Diffueraser: A diffusion model for video inpainting},
  author={Li, Xiaowen and Xue, Haolan and Ren, Peiran and Bo, Liefeng},
  journal={arXiv preprint arXiv:2501.10018},
  year={2025}
}

@inproceedings{isola2017pix2pix,
  title={Image-to-image translation with conditional adversarial networks},
  author={Isola, Phillip and Zhu, Jun-Yan and Zhou, Tinghui and Efros, Alexei A},
  booktitle={{CVPR}},
  year={2017}
}

@inproceedings{zhu2017cyclegan,
  title={Unpaired image-to-image translation using cycle-consistent adversarial networks},
  author={Zhu, Jun-Yan and Park, Taesung and Isola, Phillip and Efros, Alexei A},
  booktitle={{ICCV}},
  year={2017}
}

@inproceedings{chen2024anydoor,
  title={Anydoor: Zero-shot object-level image customization},
  author={Chen, Xi and Huang, Lianghua and Liu, Yu and Shen, Yujun and Zhao, Deli and Zhao, Hengshuang},
  booktitle={{CVPR}},
  year={2024}
}

@inproceedings{chen2025unireal,
  title={Unireal: Universal image generation and editing via learning real-world dynamics},
  author={Chen, Xi and Zhang, Zhifei and Zhang, He and Zhou, Yuqian and Kim, Soo Ye and Liu, Qing and Li, Yijun and Zhang, Jianming and Zhao, Nanxuan and Wang, Yilin and others},
  booktitle={{CVPR}},
  year={2025}
}

@inproceedings{ruiz2023dreambooth,
  title={Dreambooth: Fine tuning text-to-image diffusion models for subject-driven generation},
  author={Ruiz, Nataniel and Li, Yuanzhen and Jampani, Varun and Pritch, Yael and Rubinstein, Michael and Aberman, Kfir},
  booktitle={{CVPR}},
  year={2023}
}

@inproceedings{brooks2023instructpix2pix,
  title={Instructpix2pix: Learning to follow image editing instructions},
  author={Brooks, Tim and Holynski, Aleksander and Efros, Alexei A},
  booktitle={{CVPR}},
  year={2023}
}

@article{hertz2022prompt2prompt,
  title={Prompt-to-prompt image editing with cross attention control},
  author={Hertz, Amir and Mokady, Ron and Tenenbaum, Jay and Aberman, Kfir and Pritch, Yael and Cohen-Or, Daniel},
  journal={arXiv preprint arXiv:2208.01626},
  year={2022}
}

@article{chen2025ivebench,
  title={Ivebench: Modern benchmark suite for instruction-guided video editing assessment},
  author={Chen, Yinan and Zhang, Jiangning and Hu, Teng and Zeng, Yuxiang and Xue, Zhucun and He, Qingdong and Wang, Chengjie and Liu, Yong and Hu, Xiaobin and Yan, Shuicheng},
  journal={arXiv preprint arXiv:2510.11647},
  year={2025}
}

@inproceedings{chen2025editboard,
  title={Editboard: Towards a comprehensive evaluation benchmark for text-based video editing models},
  author={Chen, Yupeng and Chen, Penglin and Zhang, Xiaoyu and Huang, Yixian and Xie, Qian},
  booktitle={{AAAI}},
  year={2025}
}

@inproceedings{sun2025vebench,
  title={Ve-bench: subjective-aligned benchmark suite for text-driven video editing quality assessment},
  author={Sun, Shangkun and Liang, Xiaoyu and Fan, Songlin and Gao, Wenxu and Gao, Wei},
  booktitle={{AAAI}},
  year={2025}
}

@article{mou2025instructx,
  title={Instructx: Towards unified visual editing with mllm guidance},
  author={Mou, Chong and Sun, Qichao and Wu, Yanze and Zhang, Pengze and Li, Xinghui and Ye, Fulong and Zhao, Songtao and He, Qian},
  journal={arXiv preprint arXiv:2510.08485},
  year={2025}
}

@inproceedings{tan2024edtalk,
  title={Edtalk: Efficient disentanglement for emotional talking head synthesis},
  author={Tan, Shuai and Ji, Bin and Bi, Mengxiao and Pan, Ye},
  booktitle={{ECCV}},
  year={2024}
}

@inproceedings{zhang2018lpips,
  title={The unreasonable effectiveness of deep features as a perceptual metric},
  author={Zhang, Richard and Isola, Phillip and Efros, Alexei A and Shechtman, Eli and Wang, Oliver},
  booktitle={{CVPR}},
  year={2018}
}

@article{wang2004ssim,
  author  = {Zhou Wang and A. C. Bovik and H. R. Sheikh and E. P. Simoncelli},
  title   = {Image Quality Assessment: From Error Visibility to Structural Similarity},
  journal = {{TIP}},
  year    = {2004}
}

@inproceedings{psnr,
  author    = {Alain Hore and Djemel Ziou},
  title     = {Image Quality Metrics: PSNR vs. SSIM},
  booktitle = {{ICPR}},
  year      = {2010},
}

@inproceedings{huang2023vbench,
 title={{VBench}: Comprehensive Benchmark Suite for Video Generative Models},
 author={Huang, Ziqi and He, Yinan and Yu, Jiashuo and Zhang, Fan and Si, Chenyang and Jiang, Yuming and Zhang, Yuanhan and Wu, Tianxing and Jin, Qingyang and Chanpaisit, Nattapol and Wang, Yaohui and Chen, Xinyuan and Wang, Limin and Lin, Dahua and Qiao, Yu and Liu, Ziwei},
 booktitle={{CVPR}},
 year={2024}
 }

@article{he2025openve,
  title={OpenVE-3M: A Large-Scale High-Quality Dataset for Instruction-Guided Video Editing},
  author={He, Haoyang and Wang, Jie and Zhang, Jiangning and Xue, Zhucun and Bu, Xingyuan and Yang, Qiangpeng and Wen, Shilei and Xie, Lei},
  journal={arXiv preprint arXiv:2512.07826},
  year={2025}
}

@article{zhao2025risebench,
  title={Envisioning beyond the pixels: Benchmarking reasoning-informed visual editing},
  author={Zhao, Xiangyu and Zhang, Peiyuan and Tang, Kexian and Zhu, Xiaorong and Li, Hao and Chai, Wenhao and Zhang, Zicheng and Xia, Renqiu and Zhai, Guangtao and Yan, Junchi and others},
  journal={arXiv preprint arXiv:2504.02826},
  year={2025}
}

@article{liu2025revise,
  title={ReViSE: Towards Reason-Informed Video Editing in Unified Models with Self-Reflective Learning},
  author={Liu, Xinyu and Yuan, Hangjie and Wei, Yujie and Xing, Jiazheng and Han, Yujin and Pan, Jiahao and Ma, Yanbiao and Chan, Chi-Min and Zhao, Kang and Zhang, Shiwei and others},
  journal={arXiv preprint arXiv:2512.09924},
  year={2025}
}

@article{wan2025wan,
  title={Wan: Open and advanced large-scale video generative models},
  author={Wan, Team and Wang, Ang and Ai, Baole and Wen, Bin and Mao, Chaojie and Xie, Chen-Wei and Chen, Di and Yu, Feiwu and Zhao, Haiming and Yang, Jianxiao and others},
  journal={arXiv preprint arXiv:2503.20314},
  year={2025}
}

@article{seedance2026seedance,
  title={Seedance 2.0: Advancing Video Generation for World Complexity},
  author={Seedance, Team and Chen, De and Chen, Liyang and Chen, Xin and Chen, Ying and Chen, Zhuo and Chen, Zhuowei and Cheng, Feng and Cheng, Tianheng and Cheng, Yufeng and others},
  journal={arXiv preprint arXiv:2604.14148},
  year={2026}
}

@article{team2025kling,
  title={Kling-Omni Technical Report},
  author={Team, Kling and Chen, Jialu and Ci, Yuanzheng and Du, Xiangyu and Feng, Zipeng and Gai, Kun and Guo, Sainan and Han, Feng and He, Jingbin and He, Kang and others},
  journal={arXiv preprint arXiv:2512.16776},
  year={2025}
}

@article{cheng2025wananimate,
  title={Wan-animate: Unified character animation and replacement with holistic replication},
  author={Cheng, Gang and Gao, Xin and Hu, Li and Hu, Siqi and Huang, Mingyang and Ji, Chaonan and Li, Ju and Meng, Dechao and Qi, Jinwei and Qiao, Penchong and others},
  journal={arXiv preprint arXiv:2509.14055},
  year={2025}
}

@article{guo2023animatediff,
  title={Animatediff: Animate your personalized text-to-image diffusion models without specific tuning},
  author={Guo, Yuwei and Yang, Ceyuan and Rao, Anyi and Liang, Zhengyang and Wang, Yaohui and Qiao, Yu and Agrawala, Maneesh and Lin, Dahua and Dai, Bo},
  journal={arXiv preprint arXiv:2307.04725},
  year={2023}
}

@inproceedings{ho2022vdm,
  title={Video diffusion models},
  author={Ho, Jonathan and Salimans, Tim and Gritsenko, Alexey and Chan, William and Norouzi, Mohammad and Fleet, David J},
  booktitle={{NeurIPS}},
  year={2022}
}

@inproceedings{lipman2023flowmatching,
  title={Flow Matching for Generative Modeling},
  author={Lipman, Yaron and Chen, Ricky TQ and Ben-Hamu, Heli and Nickel, Maximilian and Le, Matthew},
  booktitle={{ICLR}},
  year={2023}
}

@inproceedings{saharia2022imagen,
  title={Photorealistic text-to-image diffusion models with deep language understanding},
  author={Saharia, Chitwan and Chan, William and Saxena, Saurabh and Li, Lala and Whang, Jay and Denton, Emily L and Ghasemipour, Kamyar and Gontijo Lopes, Raphael and Karagol Ayan, Burcu and Salimans, Tim and others},
  booktitle={{NeurIPS}},
  year={2022}
}

@inproceedings{peebles2023dit,
  title={Scalable diffusion models with transformers},
  author={Peebles, William and Xie, Saining},
  booktitle={{ICCV}},
  year={2023}
}

@article{labs2025flux,
  title={FLUX. 1 Kontext: Flow Matching for In-Context Image Generation and Editing in Latent Space},
  author={Labs, Black Forest and Batifol, Stephen and Blattmann, Andreas and Boesel, Frederic and Consul, Saksham and Diagne, Cyril and Dockhorn, Tim and English, Jack and English, Zion and Esser, Patrick and others},
  journal={arXiv preprint arXiv:2506.15742},
  year={2025}
}

@article{liu2022rectifiedflow,
  title={Flow straight and fast: Learning to generate and transfer data with rectified flow},
  author={Liu, Xingchao and Gong, Chengyue and Liu, Qiang},
  journal={arXiv preprint arXiv:2209.03003},
  year={2022}
}

@inproceedings{jiang2025smarteraser,
  author    = {Longtao Jiang and Zhendong Wang and Jianmin Bao and Wengang Zhou and Dongdong Chen and Lei Shi and Dong Chen and Houqiang Li},
  title     = {SmartEraser: Remove Anything from Images using Masked-Region Guidance},
  booktitle = {{CVPR}},
  year      = {2025}
}

@article{oord2016wavenet,
  title={Wavenet: A generative model for raw audio},
  author={Oord, Aaron van den and Dieleman, Sander and Zen, Heiga and Simonyan, Karen and Vinyals, Oriol and Graves, Alex and Kalchbrenner, Nal and Senior, Andrew and Kavukcuoglu, Koray},
  journal={arXiv preprint arXiv:1609.03499},
  year={2016}
}

@inproceedings{kong2020hifigan,
  title={Hifi-gan: Generative adversarial networks for efficient and high fidelity speech synthesis},
  author={Kong, Jungil and Kim, Jaehyeon and Bae, Jaekyoung},
  booktitle={{NeurIPS}},
  year={2020}
}

@inproceedings{hessel2021clipscore,
  title={Clipscore: A reference-free evaluation metric for image captioning},
  author={Hessel, Jack and Holtzman, Ari and Forbes, Maxwell and Le Bras, Ronan and Choi, Yejin},
  booktitle={{EMNLP}},
  year={2021}
}

@inproceedings{zhang2023inST,
  title={Inversion-based style transfer with diffusion models},
  author={Zhang, Yuxin and Huang, Nisha and Tang, Fan and Huang, Haibin and Ma, Chongyang and Dong, Weiming and Xu, Changsheng},
  booktitle={{CVPR}},
  year={2023}
}

@misc{xai2024grokimagine,
  author       = {xAI},
  title        = {Grok Imagine},
  year         = {2024},
  howpublished = {\url{https://x.ai}},
}

@misc{runway2024aleph,
  author       = {Runway},
  title        = {Runway Aleph},
  year         = {2024},
  howpublished = {\url{https://runwayml.com}},
}

@inproceedings{yang2026geniedrive,
  title={Geniedrive: Towards physics-aware driving world model with 4d occupancy guided video generation},
  author={Yang, Zhenya and Liu, Zhe and Lu, Yuxiang and Hou, Liping and Miao, Chenxuan and Peng, Siyi and Feng, Bailan and Bai, Xiang and Zhao, Hengshuang},
  booktitle={{CVPR}},
  year={2026}
}

@article{zeng2025lumen,
  title={Lumen: Consistent video relighting and harmonious background replacement with video generative models},
  author={Zeng, Jianshu and Liu, Yuxuan and Feng, Yutong and Miao, Chenxuan and Gao, Zixiang and Qu, Jiwang and Zhang, Jianzhang and Wang, Bin and Yuan, Kun},
  journal={arXiv preprint arXiv:2508.12945},
  year={2025}
}


\newpage
\appendix
\crefalias{section}{appendix}
\section{Data Source and Categories of OmniEdit-Bench}
The videos in \method are collected from three primary sources: (1) open-source video platforms, such as Pexels and Pixabay, which provide diverse real-world footage; (2) publicly available video datasets, including OpenVid-1M, offering large-scale and well-curated video content; and (3) state-of-the-art text-to-video generation models, which enable the inclusion of synthetic yet controllable scenarios that are difficult to obtain from real-world data.

To ensure both diversity and representativeness, we carefully curate videos spanning a wide range of categories, including plants, animals, human activities, natural environments, and human-made scenes. This diverse composition allows the benchmark to cover varied visual appearances, motion patterns, and interaction types. In addition, we explicitly balance the data across different content domains to avoid bias toward specific scene types.

\section{Prompts for VLM Evaluation}
\label{appd:prompts_for_vlm}
\textbf{Base Instruction:} You are a STRICT and PROFESSIONAL evaluator for video editing tasks. You will be given: Original Video, Edited Video, User Edit Instruction, and (Optional) Reference (image or description). Your task is to evaluate the Edited Video.Core principles:1. ACCURACY IS THE MOST IMPORTANT. If the instruction is NOT correctly followed, accuracy MUST be low (0-2). Even if the video looks realistic, incorrect instruction → low score.2. DO NOT BE FOOLED BY VISUAL QUALITY. Realism is not equal to correctness. Pretty but wrong → low score.3. PRESERVATION RULE: Only evaluate regions NOT related to the instruction.4. CONSISTENCY RULE: Check BOTH temporal and spatial stability across the entire video.5. BE STRICT: Most real-world outputs are imperfect → avoid giving 5 unless nearly perfect.

\textbf{Spatial Track:} 1) Accuracy (CRITICAL): Attribute accuracy (color, texture, size), Subject accuracy (add/remove/replace), Global accuracy (scene-level transformation). Reasoning: spatial relationships (position, layout, geometry) must be correct.2) Preservation: Unedited objects/regions must remain unchanged.3) Realism: Physical realism (lighting, shadow), Subject realism, Environment realism.4) Consistency: Temporal consistency (no flicker), Spatial consistency (structure stable).

\textbf{Temporal Track:} 1) Accuracy (CRITICAL): Action correctness, Attribute correctness (speed, freq), Timing correctness (order, delay), Camera control. Reasoning: temporal evolution must follow real-world logic.2) Preservation: Unrelated temporal content must remain unchanged.3) Realism: Motion realism (natural, smooth, no artifacts).4) Consistency: Temporal stability (no drift).

\textbf{Audio Track:} 1) Accuracy (CRITICAL): Sound correctness, Timing correctness. Reasoning: audio must match event semantics.2) Preservation: Original audio must remain when not edited.3) Realism: Audio realism (no distortion), Scene consistency.4) Consistency: Audio-visual sync (lip sync).

\textbf{Reference-based Track:} 1) Accuracy (CRITICAL): All spatial/temporal edits must follow instruction. MUST align with the provided reference (image or description). Reference alignment is REQUIRED for a high score.2) Preservation: Non-edited content should remain unchanged.3) Realism: Physical plausibility and visual fidelity.4) Consistency: Temporal and spatial consistency.

\textbf{Reasoning Track:} 1) Accuracy (CRITICAL): Explicit instruction following + Implicit reasoning correctness (Temporal/Causal/Spatial/Logical evolution). IMPORTANT: If the underlying reasoning is wrong (e.g., cause-effect logic fails), accuracy must be LOW even if some edit exists.2) Preservation: Original visual content should be preserved unless required to change.3) Realism: Physical, subject, and environment realism.4) Consistency: Temporal stability and structural integrity.

\section{Visualization Results}
\label{appd:visualization_results}
\begin{figure*}[ht!]
    \centering
    \includegraphics[width=\linewidth]{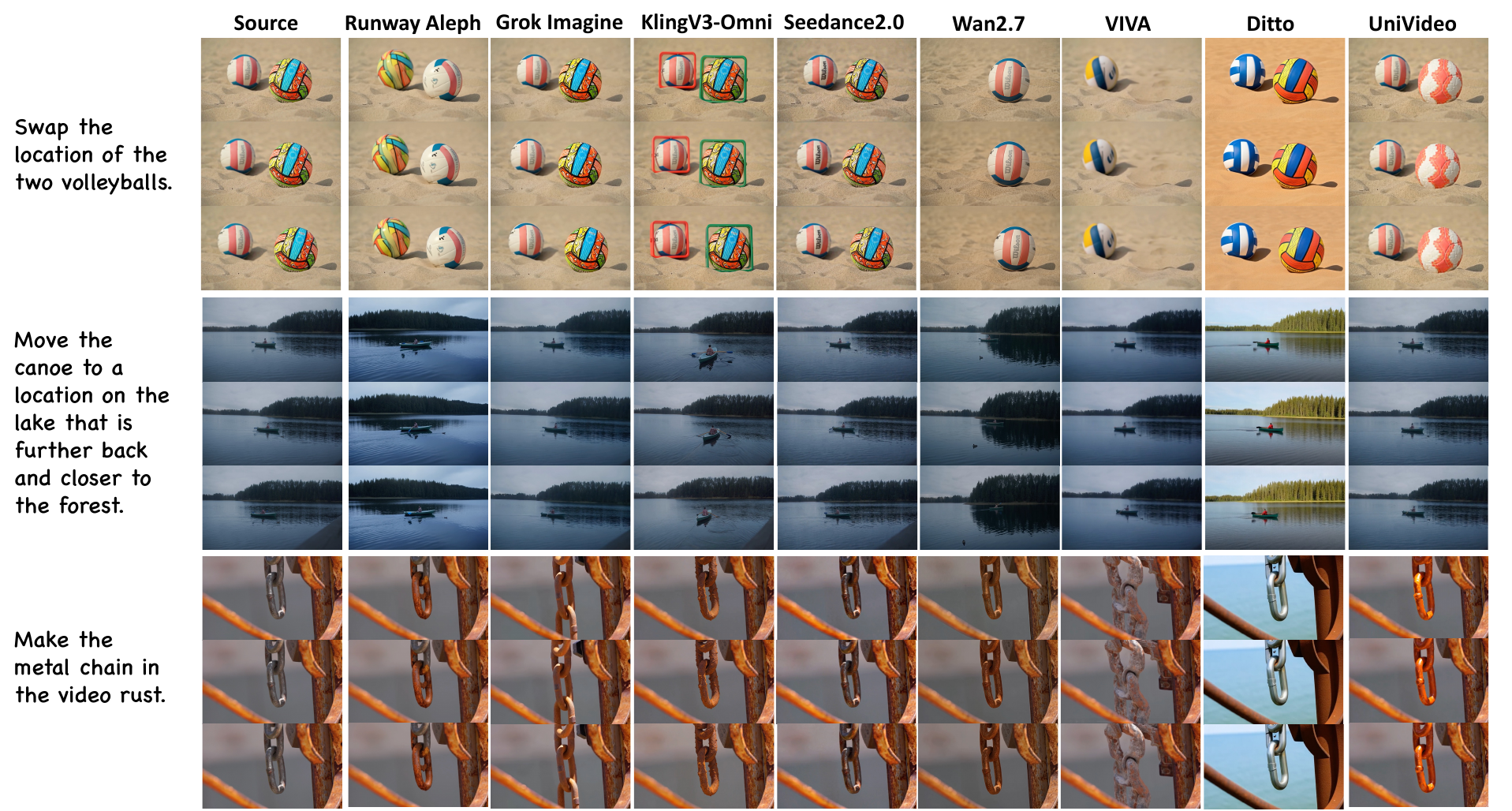}
    \caption{}
    \label{fig:visualizations_1}
\end{figure*}

\begin{figure*}[ht!]
    \centering
    \includegraphics[width=\linewidth]{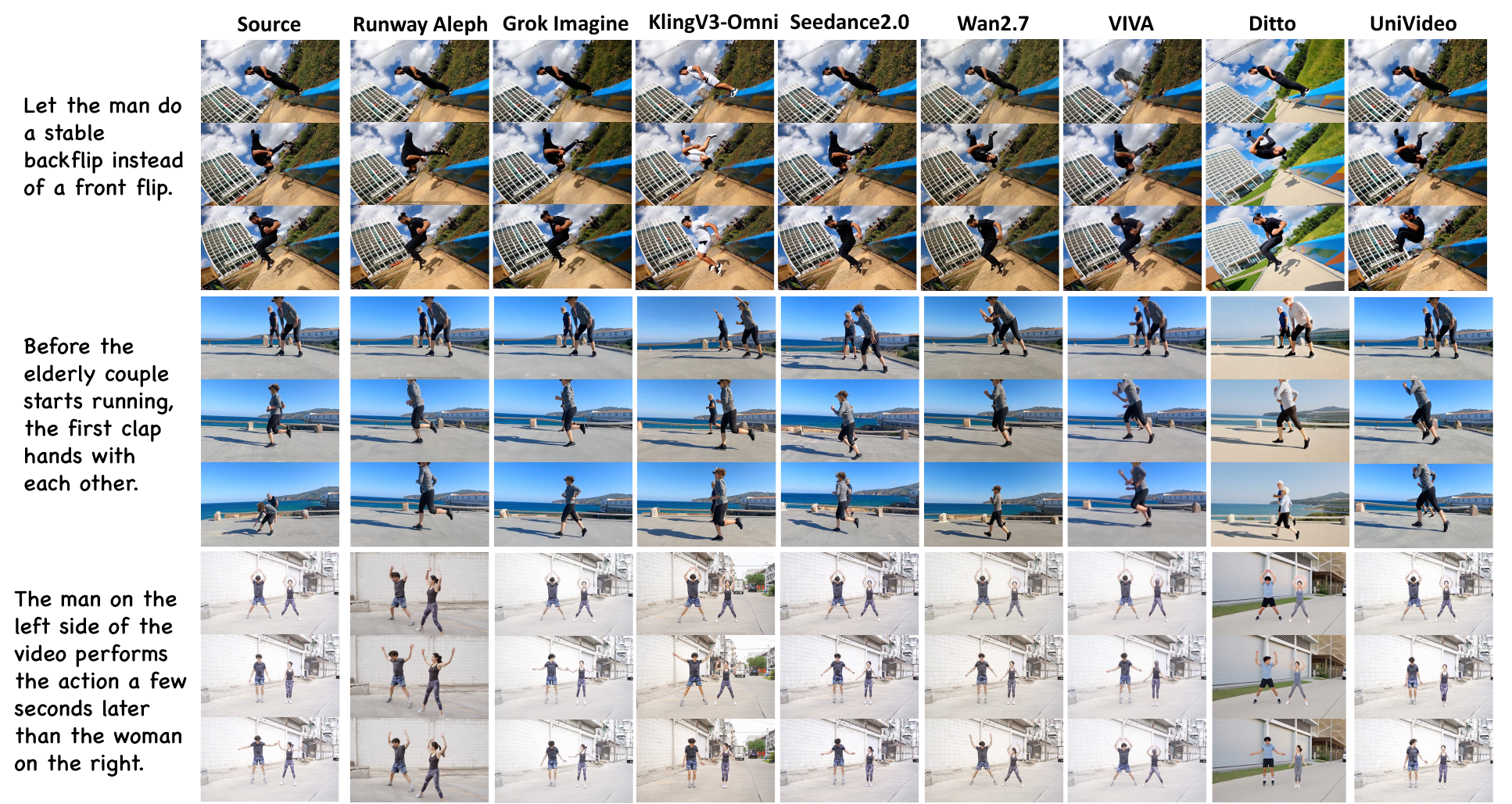}
    \caption{}
    \label{fig:visualizations_2}
\end{figure*}

\begin{figure*}[ht!]
    \centering
    \includegraphics[width=\linewidth]{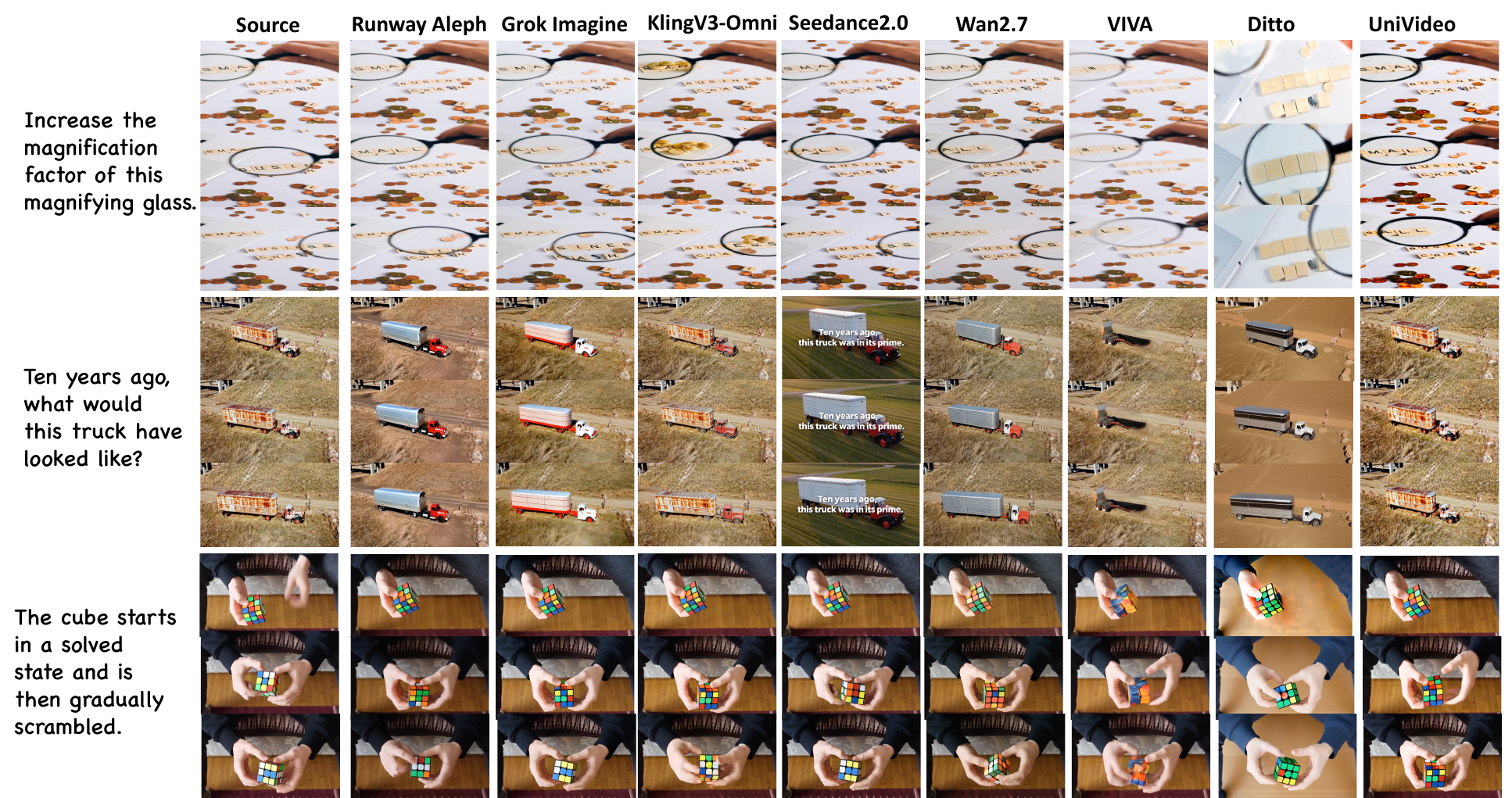}
    \caption{}
    \label{fig:visualizations_3}
\end{figure*}

We provide much more visualization results of models evaluated on our \method as shown in \cref{fig:visualizations_1}, \cref{fig:visualizations_2} and \cref{fig:visualizations_3}.

\section{Limitations}
\label{appd:limitations}
Despite its comprehensive design, \method still has several limitations. The benchmark is constrained by the scale and diversity of available data, particularly for complex temporal and audio scenarios. While VLM-based evaluation provides a scalable alternative to human annotation, it may not fully capture subtle subjective preferences or nuanced editing quality in borderline cases. In addition, the current designs of the reference-conditioned and reasoning tracks only cover a subset of possible settings, leaving more complex cross-modal dependencies and long-horizon reasoning scenarios insufficiently explored.

\section{Potential Impact}
\method provides a comprehensive benchmark for evaluating instruction-based video editing, which can facilitate the development of more controllable and reliable video editing systems. Such advances have the potential to benefit a wide range of applications, including content creation, film production, education, and accessibility, by lowering the barrier for generating and editing video content through natural language.

At the same time, improved video editing capabilities may raise concerns about misuse, particularly in generating misleading or manipulated media. The availability of more powerful editing tools could exacerbate issues such as misinformation or synthetic content abuse. By providing a standardized and interpretable evaluation framework, our work aims to promote more responsible development and assessment of video editing systems. Future work should further explore safeguards, including detection and attribution mechanisms, to mitigate potential risks.

\section{Safeguards}
To promote responsible use, \method is constructed using publicly available or properly licensed video sources, and care is taken to avoid sensitive or personal content. The benchmark is intended solely for evaluation purposes and does not provide models or tools for direct video generation or editing.

We acknowledge that advances in instruction-based video editing may raise potential risks, such as the creation of misleading or manipulated content. By focusing on standardized evaluation and analysis, our work aims to support more transparent and responsible development of video editing systems. Future efforts may further incorporate safeguards such as content filtering and misuse detection.


\newpage

\end{document}